%% file: main.tex
\documentclass[preprint]{elsarticle}

\usepackage{amssymb}
\usepackage[british]{babel}
\usepackage[misc]{ifsym}
\usepackage[acronym]{glossaries}
\usepackage{csquotes}
\usepackage{natbib}
\usepackage{verbatim}
\usepackage{amsmath}
\usepackage{multirow}
\usepackage{colortbl}
\usepackage{tikz}
\usepackage{booktabs}
\usepackage{graphicx}
\usepackage{rotating}
\usepackage[hidelinks]{hyperref}
\usepackage[normalem]{ulem}
\useunder{\uline}{\ul}{}
\usepackage[labelfont={bf}]{caption}
\usepackage{subcaption}
\usepackage{threeparttable}
\usepackage{adjustbox}
\usepackage[T1]{fontenc}
\usetikzlibrary{arrows}
\usetikzlibrary{positioning}
\usetikzlibrary{calc}
\usetikzlibrary{shapes.multipart}
\usetikzlibrary{matrix}
\usetikzlibrary{decorations.pathreplacing,arrows.meta,calligraphy}
\usetikzlibrary{decorations.text}

\newtheorem{definition}{Definition}

\definecolor{echodrk}{HTML}{0099cc}
\definecolor{olivegreen}{rgb}{0,0.6,0}
\definecolor{camdrk}{RGB}{0,62,114}

\journal{Decision Support Systems}

\input{commands.tex}

\begin{document}

\begin{frontmatter}

\title{
A Technique for Determining Relevance Scores of Process Activities using Graph-based Neural Networks
}

\author[add1]{Matthias Stierle}\corref{cor1}
\ead{matthias.stierle@fau.de}
\author[add1]{Sven Weinzierl}
\ead{sven.weinzierl@fau.de}
\author[add1]{Maximilian Harl}
\ead{maximilian.harl@fau.de}
\author[add1]{Martin Matzner}
\ead{martin.matzner@fau.de}

\cortext[cor1]{Corresponding author}

\address[add1]{Institute of Information Systems, Friedrich-Alexander-Universit\"at N\"urnberg-Erlangen, Germany}

\begin{abstract}
\input{sections/00_abstract}
\end{abstract}

\end{frontmatter}

\emph{This is the accepted manuscript of an article published by Elsevier in Decision Support Systems, available online: \url{https://doi.org/10.1016/j.dss.2021.113511}.}

\input{99_acronyms.tex}

\newpage
\input{sections/01_intro.tex}
\input{sections/02_background}
\input{sections/04_artifact.tex}
\input{sections/05_evaluation.tex}
\input{sections/06_case_study}
\input{sections/07_discussion.tex}
\input{sections/08_conclusion.tex}

\section*{Acknowledgements}
This work was supported by the German Federal Ministry of Education and Research (BMBF) within the framework programme \emph{Software Campus} (www.softwarecampus.de) [No. 01IS17045].

\bibliographystyle{model1-num-names.bst}

\bibliography{references.bib}

\end{document}

%% file: commands.tex
\newcommand{\magic}[0]{\textit{\gls{magic}}} %
\newcommand{\serviceprocessdata}[0]{sp2020} %

%% file: sections/00_abstract.tex
Process models generated through process mining depict the as-is state of a process. Through annotations with metrics such as the frequency or duration of activities, these models provide generic information to the process analyst. To improve business processes with respect to performance measures, process analysts require further guidance from the process model. In this study, we design \emph{Graph Relevance Miner (GRM)}, a technique based on graph neural networks, to determine the relevance scores for process activities with respect to performance measures. Annotating process models with such relevance scores facilitates a problem-focused analysis of the business process, placing these problems at the centre of the analysis. We quantitatively evaluate the predictive quality of our technique using four datasets from different domains, to demonstrate the faithfulness of the relevance scores. Furthermore, we present the results of a case study, which highlight the utility of the technique for organisations. Our work has important implications both for research and business applications, because process model-based analyses feature shortcomings that need to be urgently addressed to realise successful process mining at an enterprise level.

%% file: 99_acronyms.tex
\newacronym[\glslongpluralkey={decision support systems}]{dss}{DSS}{decision support system}
\newacronym{pbpm}{PBPM}{predictive business process monitoring}
\newacronym{bpm}{BPM}{business process management}
\newacronym{ml}{ML}{machine learning}
\newacronym{ai}{AI}{artificial intelligence}
\newacronym{xai}{XAI}{explainable artificial intelligence}
\newacronym{gnn}{GNN}{graph-based neural network}
\newacronym{ggnn}{GGNN}{gated graph neural network}
\newacronym{mpnn}{MPNN}{message passing neural network}
\newacronym{gru}{GRU}{gated recurrent unit}
\newacronym{pig}{IG}{instance graph}
\newacronym{magic}{GRM}{Graph Relevance Miner}
\newacronym{dfg}{DFG}{directly-follows graph}
\newacronym{dsr}{DSR}{design science research}
\newacronym{bilstm}{BiLSTM}{bi-directional long short-term memory DNN}
\newacronym{dl}{DL}{deep learning}
\newacronym{dnn}{DNN}{deep neural network}
\newacronym{kpi}{KPI}{key performance indicator}
\newacronym{ppi}{PPI}{process performance indicator}
\newacronym{pm}{PM}{process mining}

%% file: sections/01_intro.tex
\section{Introduction}
\label{sec:intro}
The purpose of \gls{bpm} is to improve business processes~\citep{VanDerAalst2016}. 
A central role in process improvement is played by the process analyst \citep{Rosemann2010}, who is responsible for \textquote{monitoring, measuring, and providing feedback on the performance of a business process}~\citep[p.45]{Sonteya2012}. 
The ongoing implementation of information systems in organisations, along with the subsequently enhanced availability of event log data, have enabled process analysts to discover as-is models of processes with \gls{pm} with relative ease \citep{van2018process}. 
However, the crucial challenge lies in identifying potential areas for process improvements (i.e., process analysis) with respect to a strategic goal \citep{beverungen2020seven}; this requires analytical capabilities such as Pareto or root cause analysis \citep{Rosemann2010}.

A business process can be defined as a \textquote{completely closed, timely, and logical sequence of activities}~\citep[p.3]{Becker2003} that realises an outcome valuable to a customer \citep{Dumas2013}.
The effectiveness (i.e., customer value) and efficiency (e.g., timely, logical sequence, resource utilisation) of a business process are monitored using \glspl{kpi} as aggregated measures of process outcomes; in the context of \gls{bpm}, these are often referred to as \glspl{ppi} \citep{DELRIOORTEGA2013470}. %
Thus, to improve a business process, it is essential for a process analyst to understand the relevance of individual process activities in terms of their impact on the dimensions expressed by these performance measures.

For example, we consider a travel reimbursement process at an university; it aims for a high degree of compliance with travel policies. Observing that the \gls{kpi} \emph{ratio of budget violations} increases, the process analyst must understand which activities in the process should be redesigned to improve the process; hence, they must evaluate the \gls{kpi}. 
In Figure~\ref{fig:example_discovery_bpi2020}, two discovered process models are presented. On the left, the most frequent path is shown, annotated with the number of occurrences for each activity. The process analyst can deduce information about the reimbursement process’ execution from a \emph{generic} perspective but not with respect to the budget violations. The right-hand process model indicates the most relevant path in terms of budget violations, and each process activity is annotated with a relevance score expressing its importance thereto. The process analyst can directly identify which activities should be considered for redesign and can also suggest which of these activities should be considered first (e.g., \emph{Permit A}). 

\input{img/fig_example_processdiscovery_bpi2020}

Process model-based analysis---that is, process analysis based on the discovered process model---is able to make users aware of the business processes behind the data and can subsequently guide process analysts as they improve these processes \citep{vanEck2019}. To facilitate analysis beyond the simple discovery of a process, the process model must provide information suitable for the improvement initiative.
Therefore, we design a technique to determine the relevance scores of process activities with respect to a performance measure extracted from the event log data.

Determining relevance scores for process activities is an interesting challenge, owing to the plurality of relationships between activities. 
For instance, an activity may or may not occur; if it does, then it may occur towards the start or end of a process, once or multiple times, and before, after, or between other activities, etc.  
Understanding these complex relationships and their influences on process performance is a difficult task.

To address this challenge, the field of \gls{ml} provides a set of algorithms that automatically discover structures in data and capture those as mathematical models representing functions \cite{bishop2006pattern}.     
In the context of \gls{ml}, \glspl{dnn} have \textquote{turned out to be very good at discovering the intricate structures in high-dimensional data and is therefore applicable to many domains}~\citep[p.436]{lecun2015deeplearning}.
Among others, \citet{evermann2017}, \citet{mehdiyev2020novel}, and \citet{Kratsch2020} showed that with \gls{dl}, in particular \glspl{dnn}, predictive models can be learned from event log data more accurately than with traditional \gls{ml} techniques. 
However, \glspl{dnn} often struggle to intuitively represent the learned structures; this is commonly referred to as the \emph{black-box} problem \citep{castelvecchi2016}.

\Glspl{gnn} are a relatively new group of \glspl{dnn}; they have proved useful in domains where the input data have a graph structure, such as in chemistry and molecular biology \citep{scarselli2008graph}.
Compared to \emph{traditional} \glspl{dnn} such as multi-layer perceptrons, \glspl{gnn} can compute graph data directly~\cite{zhou.2018}. 
In particular, the structure of the input graph can be matched directly to the topology of the \gls{gnn}; this allows for direct inferences to be made between the relevance of network nodes and graph nodes.
\Glspl{ggnn} are a variant of \glspl{gnn}; they were designed to tackle temporal dependencies in the data~\citep{li2015gated}; such dependencies are a significant aspect of event log data.

The main idea of this paper is to design a \gls{ggnn}-based technique---referred to as \magic{}---to determine the relevance scores (with respect to a given performance measure) of process activities from event log data. 
First, we transform process instances using a prediction label (i.e., the performance measure), converting them into \glspl{pig}. 
Second, we input these graphs into the \gls{ggnn} model for training and testing. Finally, we input multiple instances into the \gls{ggnn} model, to determine the relevance scores.

This paper builds on previous work where we demonstrated the use of \glspl{gnn} with process data \citep{harl2020ggnn} within the context of \emph{explainable artificial intelligence}. We presented a technique that provides explanations for predictions made for the outcome of single process instances to process workers. In this paper, we follow up on the idea to extract relevance scores from \gls{gnn}. However, we shift the focus from the prediction of single instances for process workers onto providing insights on the process level for process analysts. We formally describe the technique and evaluate both its efficacy and its effectiveness with empirical data \citep{march1995design}.

The remainder of this paper is organised as follows. In Section \ref{sec:background}, we introduce the preliminary information regarding event logs and \glspl{gnn}. Next, we present the design of our technique in Section \ref{sec:artifact}. In Section \ref{sec:eval}, we present the results from our evaluation of the technique, obtained for four different real-life datasets; then, we describe our case study. 
In Section \ref{sec:discussion}, we summarise our contributions, review the related works, and consider the limitations of our study. Lastly, in Section \ref{sec:conclusion}, we conclude the paper with a brief summary of the technique’s potential impacts on research and business applications, and we highlight possible future research directions.

%% file: img/fig_example_processdiscovery_bpi2020.tex
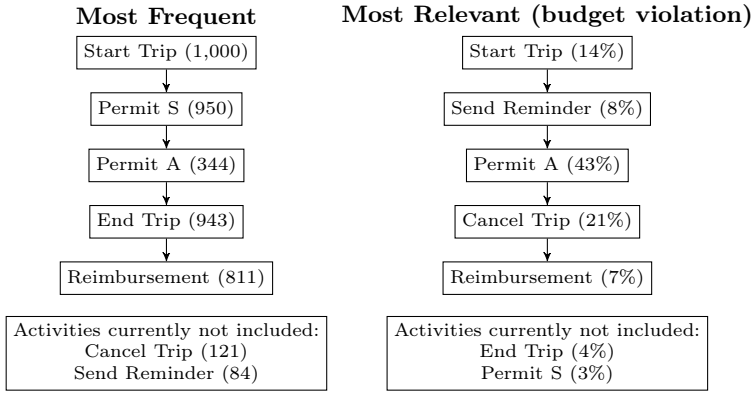
\begin{figure}[htb]
    \centering
    
\scriptsize
\begin{tikzpicture}[->,>=stealth',auto,node distance=3cm, main node/.style={circle,draw},every text node part/.style={align=center}]
\tikzset{
    vertex/.style={minimum size=1.5em},
    edge/.style={->,> = latex'}
}

\end{tikzpicture}

\begin{tikzpicture}[->,>=stealth',auto,node distance=3cm, main node/.style={circle,draw},every text node part/.style={align=center}]
\tikzset{
    vertex/.style={minimum size=1.5em},
    edge/.style={->,> = latex'}
}

\node[vertex, draw] (1) at (0,0) {Start Trip (1,000)};
\node[vertex, draw] (2) at (0,-0.75) {Permit S (950)};
\node[vertex, draw] (3) at (0,-1.5) {Permit A (344)};
\node[vertex, draw] (4) at (0,-2.25) {End Trip (943)};
\node[vertex, draw] (5) at (0,-3)  {Reimbursement (811)};

\path[every node/.style={font=\sffamily\small,
  		fill=white,inner sep=1pt}]
  	(1) edge  node {} (2)
    (2) edge  node {} (3)
    (3) edge  node {} (4)
    (4) edge  node {} (5);

\node[draw,align=left] at (0,-4) {Activities currently not included:\\ Cancel Trip (121) \\ Send Reminder (84)};
\node[above,font=\small\bfseries] at (current bounding box.north) {Most Frequent};

\end{tikzpicture}
\begin{tikzpicture}[->,>=stealth',auto,node distance=3cm, main node/.style={circle,draw},every text node part/.style={align=center}]
\tikzset{
    vertex/.style={minimum size=1.5em},
    edge/.style={->,> = latex'}
}

\node[vertex, draw] (1) at (0,0) {Start Trip (14\%)};
\node[vertex, draw] (2) at (0,-0.75) {Send Reminder (8\%)};
\node[vertex, draw] (3) at (0,-1.5) {Permit A (43\%)};
\node[vertex, draw] (4) at (0,-2.25) {Cancel Trip (21\%)};
\node[vertex, draw] (5) at (0,-3)  {Reimbursement (7\%)};

\path[every node/.style={font=\sffamily\small,
  		fill=white,inner sep=1pt}]
  	(1) edge  node {} (2)
    (2) edge  node {} (3)
    (3) edge  node {} (4)
    (4) edge  node {} (5);

\node[draw,align=left] at (0, -4) {Activities currently not included:\\ End Trip (4\%) \\ Permit S (3\%)};
\node[above,font=\small\bfseries] at (current bounding box.north) {Most Relevant (budget violation)};
\end{tikzpicture}
    \caption{Process visualisation: frequency vs. relevance.}
    \label{fig:example_discovery_bpi2020}
\end{figure}

%% file: sections/02_background.tex
\section{Background}
\label{sec:background}

\subsection{Event Logs}

\Gls{pm} is a technology that facilitates the discovery, analysis, and enhancement of process models, using the data extracted from event logs \citep{van2018process}. 
An event log is structured into traces, which are in turn structured into events. Thus, based on \citet{polato.2014}, we define the terms \textit{event universe}, \textit{event}, \textit{trace}, and \textit{event log} as follows: 

\begin{definition}[Event universe] $\mathcal{E}= \mathcal{A} \times C \times \mathcal{T}$ is the event universe in which $\mathcal{A}$ is the set of process activities, $\mathcal{C}$ the set of process instances~(cases), $C$ the set of case IDs with the bijective projection $id : C \to \mathcal{C}$, and $\mathcal{T}$ the set of timestamps.
To consider time, a process instance $c \in \mathcal{C}$ contains all past and future events, whereas events in the trace $\sigma_c$ of $c$ contain all events up to the current time instant.  
\end{definition}

\begin{definition}[Event] 
An event $e \in \mathcal{E}$
  is a tuple $e=(a,c,t)$,
  where $a \in \mathcal{A}$
  is the process activity,
  $c \in C$ is the case ID, and 
  $t \in \mathcal{T}$ is its starting timestamp.
  Given an event $e$,
  we define the projection functions
  $F_{p}=\{f_{a}, f_{c}, f_{t}\}$: $f_{a}: e \to a, f_{c}: e \to c$, and $f_{t}: e \to t$.
\end{definition}  

\begin{definition}[Trace]
A trace is a non-empty sequence $\sigma_{c} = \langle e_{1}, \dots, e_{\vert \sigma_{c} \vert} \rangle \in \mathcal{E}^*$ of events, such that $f_{c}(e_{i}) = f_{c}(e_{j}) \wedge f_{t}(e_i) \leq f_{t}(e_j)$ for $1 \leq i < j \leq \vert \sigma_{c} \vert$.
A trace can also be considered as a sequence of vectors, in which a vector contains all or part of the information relating to an event (e.g., an event's activity). Formally, $\sigma=\left\langle\mathbf{x}^{(1)}, \mathbf{x}^{(2)}, \ldots, \mathbf{x}^{(t)}\right\rangle,$ where $\mathbf{x}^{(i)} \in \mathbb{R}^{{n} \times 1}$ is a vector, and the superscript denotes the time-ordering of the events.
\end{definition}

\begin{definition}[Event log]
An event log $\mathcal{L}_\tau$
  for time instant $\tau$
  is the set of traces
  such that $\forall \sigma{}_c \in \mathcal{L}_\tau\,$,
  $\exists c \in \mathcal{C}\,$ with
  $\, \forall e \in \sigma_c$ $.$
    $id(f_c(e)) = c \wedge \forall e \in \sigma_c$ $.$ $f_t(e) \leq \tau$
  (i.e., all events of the observed cases that have already happened).
\end{definition}

Finally, our technique assumes a labelled event log for training. Thus, we define the term \textit{label}.  

\begin{definition}[Label] Given a trace $\sigma=\left\langle e_{1},\dots, e_{k}, \dots, e_{\vert \sigma \vert }\right\rangle$, we can define its label as $f_{l}(\sigma)= l$. In this paper, a label represents a certain outcome of a process; for example,\enquote{loan is accepted} or \enquote{loan is not accepted} in the case of a loan application process.
\end{definition}

\subsection{Graph Neural Networks}
\Glspl{gnn}~\cite{scarselli2008graph} are a type of neural network in which the network architecture is defined according to a graph structure. Because graphs constitute an integral part of these neural networks, we define the term \emph{graph} first. 
\begin{definition}[Graph] A tuple $G =(V,E)$ is a graph, where $V$ is a set of nodes and $E$ a set of edges. A node $v\in V$ has a unique value assigned to it, whilst an edge is a pair $\mathring{e}=\left(v, v^{\prime}\right) \in V \times V$. 
The node vector (node representation or node embedding) for node $v$ is denoted by $\mathbf{h}_{v} \in \mathbb{R}^{D}$. $D$ denotes the vector dimensionality of node $v$. 
Graphs can also contain node labels $\mathbf{l}_{v} \in\left\{\mathbf{l}_{1}, \ldots, \mathbf{l}_{\vert V \vert}\right\}$ for each node $v$, as well as edge labels $\mathbf{l}_{e} \in\left\{\mathbf{l}_{1}, \ldots, \mathbf{l}_{\vert E \vert}\right\}$ for each edge.
\end{definition}

Furthermore, we define four functions to help us manage these graphs.

\begin{definition}[Graph functions]
$f_{in}(v)=\left\{v^{\prime} \mid\left(v^{\prime}, v\right) \in E\right\}$ returns the set of
predecessor nodes $v^{\prime}$, with $v^{\prime} \rightarrow v$.
$f_{out}(v)=\left\{v^{\prime} \mid\left(v, v^{\prime}\right) \in E\right\}$ returns the set of successor nodes $v^{\prime}$, with edges $v \rightarrow v^{\prime}.$ 
$f_{nbr}(v)=f_{in}(v) \cup f_{out}(v)$ returns the set of all nodes neighbouring $v$.
$f_{co}(v)=\left\{\left(v^{\prime}, v^{\prime \prime}\right) \in E \mid v=v^{\prime} \vee v=v^{\prime \prime}\right\}$ is the set of all edges going into or out of $v$.
\end{definition}

\Glspl{gnn} map graphs to outputs via two steps. First, a propagation model computes the node representations $\mathbf{h}_v$ for each node $v$. 
Through this, the model propagates node representations over time.
The initial node representations $\mathbf{h}_{v}^{(0)}$ are set to arbitrary values.
Then, until convergence is reached, each node representation $\mathbf{h}_v^{(t+1)}$ is updated according to a \textit{local transition} function $f_{lt}$:
\begin{equation}
\label{eq:gnn0}
    \mathbf{h}^{(t+1)}_v = f_{lt}(\textbf{l}_v, \textbf{l}_{f_{co(v)}}, \mathbf{h}^{(t)}_{f_{nbr(v)}}, \textbf{l}_{f_{nbr(v)}}).
\end{equation}
The recurrent function $f_{lt}$ is shared among all nodes. Its input parameters are as follows: $\mathbf{l}_{v}$ (features of node $v$), $\mathbf{l}_{f_{co(v)}}$ (features of node $v$'s edges), $\mathbf{h}_{f_{nbr}(v)}$ (states of the neighbouring nodes; i.e., nodes that are directly connected) and $\mathbf{l}_{f_{nbr(v)}}$ (features of the neighbouring nodes). 
For this, \citet{scarselli2008graph} have suggested decomposing $f_{lt}(\cdot)$ into a sum of terms describing ingoing and outgoing edges:
\begin{equation}
\begin{split}
\label{eq:gnn1}
    \mathbf{h}^{(t+1)}_v =
    &\sum_{v^{\prime}\in f_{in}(v)}f_{lt}(\mathbf{l}_{v}, \mathbf{l}_{(v^{\prime}, v)}, \mathbf{l}_{v^{\prime}}, \mathbf{h}_{v^{\prime}}^{(t)})\text{ } + \\
    &\sum_{v^{\prime}\in f_{out}(v)}f_{lt}(\mathbf{l}_{v}, \mathbf{l}_{(v, v^{\prime})}, \mathbf{l}_{v^{\prime}}, \mathbf{h}_{v^{\prime}}^{(t)}),
\end{split}    
\end{equation}
where $f_{lt}$ is either a feed-forward neural network or a linear function of $\mathbf{h}_{v^{\prime}}$. The terms' parameters differ according to the label configuration (i.e., $\mathbf{l}_{(v^{\prime}, v)}$ or $\mathbf{l}_{(v, v^{\prime})}$, where each vector represent edge type and direction). For example, in the linear case, $f_{lt}$ can be defined as follows:
\begin{equation}
\label{eq:gnn2}
    f_{lt}(\mathbf{l}_{v}, \mathbf{l}_{(v^{\prime}, v)}, \mathbf{l}_{v^{\prime}}, \mathbf{h}_{v^{\prime}}^{(t)}) = \mathbf{A}^{(\mathbf{l}_v, \mathbf{l}_{(v^{\prime}, v),}, \mathbf{l}_v^{\prime})} \mathbf{h}_{v^{\prime}}^{(t)} + 
    \mathbf{b}^{(\mathbf{l}_v, \mathbf{l}_{(v^{\prime}, v),}, \mathbf{l}_v^{\prime})}, 
\end{equation}
where $\mathbf{A}^{(\mathbf{l}_v, \mathbf{l}_{(v^{\prime}, v),}, \mathbf{l}_v^{\prime})}$ is the sparsity matrix (or adjacency matrix) containing the weight of the edge running from node $v^{\prime}$ to $v$, and $\mathbf{b}^{(\mathbf{l}_v, \mathbf{l}_{(v^{\prime}, v),}, \mathbf{l}_v^{\prime})}$ is the bias of this edge. Both the weight and bias are learnable parameters.

After computing node representations using the propagation model, the output model maps these representations and their corresponding labels to an output. Depending on the problem to be addressed, the output can be graph-based, node-based, or edge-based. In this work, we focus on graph-based outputs, because the outcome of a process is not determined by a single node or edge.  
The graph-based output $\hat{\mathbf{o}}$ is calculated by a \textit{local output} function $f_{lo}$:  
\begin{equation}
\label{eq:gnn3}
    \hat{\mathbf{o}} = f_{lo}(\mathbf{h}^{(T)}, \mathbf{h}^{0}).
\end{equation}

Similar to the function $f_{lt}$, $f_{lo}$ is either a feed-forward neural network or a linear function of $\mathbf{h}_{v}$. To summarise, the computations described in $f_{lt}$ and $f_{lo}$ can be interpreted as feed-forward neural networks or linear functions, and their (internal) parameters are updated through a gradient-descent strategy.

Lastly, we adopt a framework for standardising \glspl{gnn}---referred to as message passing neural network (MPNN)~\citep{gilmer2017neural}---to provide a more intuitive understanding of GNN’s operation.
Corresponding to the MPNN framework, a \gls{gnn}'s propagation and output step---as defined in~\citet{scarselli2008graph}---can be described in terms of message-passing and readout phases. %
In contrast to the propagation phase, the message-passing phase updates the node representations $\mathbf{h}_v$ of node $v$ over $T$ time steps, by using messages $\mathbf{m}_v$.
Node $v$'s messages $\mathbf{m}_v$ are calculated from its neighbourhood $f_{nbr(v)}$ via the message function $M_{t}$, as 
\begin{equation}
\label{eq:mp1}
    \textbf{m}_{v}^{(t+1)} = \sum_{w\in f_{nbr(v)}} M_{t}(\textbf{h}_v^{(t)}, \textbf{h}_w^{(t)}, \mathbf{e}_{(v,w)}),
\end{equation}
where $\mathbf{h}_{v}^{(t)}$ and $\mathbf{h}_{w}^{(t)}$ are the node representations of nodes $v$ and $w$, respectively; $\mathbf{e}_{(v,w)}$ represents the features of the edge running from node $v$ to node~$w$.
Then, a node update function $U_{t}$ calculates node $v$'s new node representation $\mathbf{h}_{v}^{(t+1)}$, as formalised in Eq. (\ref{eq:mp2}):

\begin{equation}
\label{eq:mp2}    
    \textbf{h}_{v}^{(t+1)} = U_{t}(\textbf{h}_v^{(t)}, \textbf{m}_{v}^{(t+1})).
\end{equation}
Second, the readout phase uses function $R$ to omit the input parameter $\mathbf{h}^{0}$ of the GNN output function $f_{lo}$, as formalised in Eq. (\ref{eq:rf}):
\begin{equation}
\label{eq:rf}
    \hat{\mathbf{y}} = R(\{\textbf{h}_v^{(T)}| v \in G\}).
\end{equation}

\subsection{Gated Recurrent Units of the Gated Graph Neural Network}
In this paper, we adopt the GGNN architecture described in~\citet{li2015gated}. This architecture extends the \textquote{vanilla} GNN of \citet{scarselli2008graph}, using \glspl{gru}.
A GRU~\citep{cho2014learning} can be considered as a logical unit employing two gates to control the information flow over time. These two gates are referred to as reset and forget gates.
The reset gate determines the quantity of past information (from previous time steps) to be forgotten; conversely, the update gate determines the quantity of past information to be propagated to the future.
Given a sequence of inputs, a GRU computes the sequence of activations via the following recurrent equations:
\begin{equation}
    \label{eq:gru0}
    \textbf{z}^{(t)} = \text{sig}(\mathbf{W}^{T}_{z} \mathbf{x}^{(t)} + \mathbf{U}_z \textbf{h}^{(t-1)} + \mathbf{b}_{z}),
\end{equation}
\begin{equation}
    \mathbf{r}^{(t)} = \text{sig}(\mathbf{W}^{T}_r \mathbf{x}^{(t)} + \mathbf{U}_r \textbf{h}^{(t-1)} + \mathbf{b}_{r}),
\end{equation}
\begin{equation}
    \tilde{\mathbf{h}}^{(t)} =  \text{tanh}(\mathbf{W}_{h}^{T}\mathbf{x}^{(t)} + \mathbf{U}_{h}^{T}(\mathbf{r}^{(t)} \odot \mathbf{h}^{(t-1)}) + \mathbf{b}_{h}),
\end{equation}
\begin{equation}
    \label{eq:gru3}
    \textbf{h}^{(t)} = \mathbf{z}^{(t)} \odot \mathbf{h}^{(t-1)} + (1-\mathbf{z}^{(t)}) \odot \tilde{\mathbf{h}}^{(t)},
\end{equation}
where sig denotes the sigmoid activation function, $\mathbf{r}$ is the reset gate vector, $\mathbf{z}$ is the update gate vector, $\odot$ indicates a point-wise multiplication, $\mathbf{h}$ is a hidden state vector, $\mathbf{b}$ is a bias vector, and $\mathbf{W}$ and $\mathbf{U}$ are weight matrices. 
To summarise, the set $\theta =\{ \mathbf{W}, \mathbf{U}, \mathbf{b}\}$ includes the GRU's learnable parameters (i.e., its weights and biases).
Finally, we define the projection function $f_{GRU}: (\mathbf{h}^{(t)}, \mathbf{x}^{(t)}) \to \mathbf{h}^{(t+1)}$, which applies Eqs. (\ref{eq:gru0}) to~(\ref{eq:gru3}).

%% file: sections/04_artifact.tex
\section{\magic{} -- Determining Relevance Scores of Process Activities with GGNNs}
\label{sec:artifact}
\begin{figure}[htb]
\centering
  \includegraphics[width=.9\textwidth]{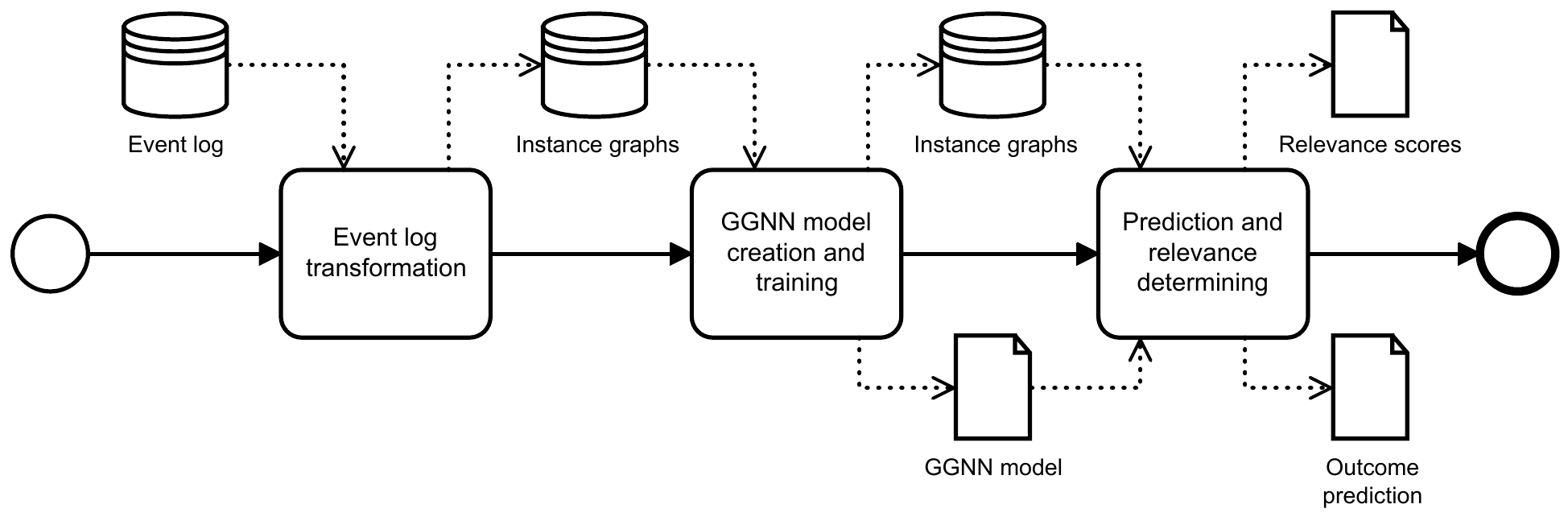}
  \caption{Our three-step GGNN-based technique for determining activity relevance scores.}
  \label{fig:artifact}
\end{figure}

\magic{} determines the relevance scores for activities using the event log data. \magic{} is based on \glspl{ggnn}. 
\magic{} consists of three steps:
\textit{(1) event log transformation}, 
\textit{(2) GGNN model creation and training}, and 
\textit{(3) prediction and relevance determining}. The steps are depicted in Figure~\ref{fig:artifact}.

First,~\magic{} loads an event log, transforms it into \glspl{pig}, and numerically encodes the \glspl{pig}' categorical values. 
To determine process activity relevance scores, a GGNN model requires a graph-oriented representation of the event log data, in the form of \glspl{pig}. 
Second, \magic{} receives as its input the \glspl{pig} from the previous step, creating and training the GGNN model therefrom. 
In the last step, \magic{} feeds the \glspl{pig} into the GGNN model, to calculate the outcome prediction.
Thus, it determines the relevance scores for activities, using \glspl{pig} from the GGNN model.

In the following, we refer to the representative event log $\mathcal{L}_\tau^{ex}$---as depicted in Table~\ref{tab:log}---to describe our technique's steps.
$\mathcal{L}_\tau^{ex}$ comprises the trace $\sigma_{1}$, which represents \textit{Case} $1$ of a reimbursement process for business travel\footnote{Note: This case originates from the \textit{bpi2020pl} event log, which is introduced in Section~\ref{sec:logs}.}.
Along with the three control-flow attributes \textit{Case}, \textit{Activity}, and \textit{Timestamp}, the event log includes the data attribute \textit{Travel expense overspent}.
The data attribute takes either the value \textquote{true} or \textquote{false}; this indicates whether or not the travel expense was excessive.
We consider this data attribute as the target attribute for learning, and we use it to apply our technique's GGNN model~$\mathcal{M}$. 
\input{tables/tab_eventlog}
\subsection{Event Log Transformation}
First, our technique transforms the event log data into a graph-oriented representation that can be employed by the GGNN model~$\mathcal{M}$. 
The transformation procedure consists of three steps: (1) event log importing, (2) \gls{pig} creation, and (3) numerical encoding.

To begin, \magic{} loads an event log $\mathcal{L}_\tau$. 
This event log $\mathcal{L}_\tau$ is transformed into a data set $\mathcal{D}$, where each instance represents a sequence of activities. As previously described, the activity and timestamp constitute elements of an event tuple. The events of the sequence are sorted by their timestamp values.
To obtain the sequence of activities of trace $\sigma_{c}$ from $\mathcal{L}_\tau$, we use the projection function $f_a(e)$ $\forall e \in \sigma_{c}$. Moreover, the target attribute's value for each case is stored in a global label vector.
After transformation, the trace $\sigma_{1}$ of the event log $\mathcal{D}$ is represented, as shown in~(\ref{ex:raw}).
\begin{equation}
\label{ex:raw}
\begin{split}
    \sigma_1 = \langle 
    &\langle \text{Start Trip}\rangle,
    \langle \text{Permit S}\rangle,
    \langle \text{Permit A}\rangle,\\
    &\langle \text{Permit A}\rangle,
    \langle \text{Permit F\_A}\rangle,
    \langle \text{End trip}\rangle,\\
    &\langle \text{Send Reminder}\rangle,
    \langle \text{Send Reminder}\rangle 
    \rangle.
\end{split}
\end{equation}
Second, \magic{} transforms the dataset $\mathcal{D}$ into a set of \glspl{pig} $\mathcal{I}$.
To this end, \citet{vandongen.2004} and \citet{diamantini2016building} have proposed methods to map sequences of events (i.e., traces) onto directed graphs of events, to enhance the transparency of the event log's traces in an isolated or aggregated manner.    
In these methods, the node of a graph instance represents an event.
However, we are here interested in the relevance scores of activities (i.e., event types) on the prediction outcome.
Thus, we introduce a definition of the \gls{pig}, in which a node denotes an activity.  
\begin{definition}[Process instance graph]
Given the trace $\sigma_c$ representing a sequence of activities for dataset $\mathcal{D}$, an \gls{pig} is a tuple of two elements $\Psi_{\sigma_c} = (V^{\Psi}_{\sigma_c}, E^{\Psi}_{\sigma_c})$, where $V^{\Psi}_{\sigma_c}$ denotes the set of nodes extracted from the trace $\sigma_{c}$, and $E^{\Psi}_{\sigma_c}$ denotes the set of edges extracted from the trace $\sigma_{c}$.
For an activity $a \in \sigma_c$, we define the projection function $f_{v}: a \to v$; $\forall a \in \sigma_{c}$, we apply $f_{v}(.)$ to obtain $V^{\Psi}_{\sigma_c}$. Hence, each activity $a \in \sigma_{c}$ is mapped to a node $v \in V^{\Psi}_{\sigma_{c}}$.
Furthermore, we add the \textquote{pseudo}-activity $\langle Start/End\rangle$ in form of a node to the set of nodes $V^{\Psi}_{\sigma_c}$.
An edge $\mathring{e}$ connecting two nodes is represented by a tuple of two temporally ordered nodes $( f_{v}(a_{i}), f_{v}(a_{j})) \in \sigma_{c}$, with $0 < i \leq j \leq \vert \sigma_{c} \vert$.    
Moreover, we add two edges to the set of edges $E^{\Psi}_{\sigma_c}$. First, the edge $(f_{v}(\langle Start/End\rangle), f_{v}(a_{1}))$ from node $f_v(\langle Start/End\rangle)$ to node $f_{v}(a_{1})$ represents the first activity of $\sigma_{c}$. 
Second, the edge $(f_{v}(a_{\vert \sigma_{c} \vert}), f_{v}(\langle Start/End\rangle))$ from node $f_{v}(a_{\vert \sigma_{c} \vert})$ to node $f_v(\langle Start/End\rangle)$ represents the last activity of $\sigma_{c}$. 
\end{definition}
\input{img/fig_example_pig}

We introduce the activity \textquote{Start/End} as a node in $V^{\Psi}_{\sigma_{c}}$ of $\Psi_{\sigma_{c}}$, to indicate the start and end of the original trace $\sigma_{c}$. 
The GGNN model $\mathcal{M}$ expects the \glspl{pig} for such a \textquote{Start/End} activity to preserve the correct ordering of the instances' activities in the model-learning and prediction phases.
For example, the trace $\sigma_{1}$ from Eq. (\ref{ex:raw}) is transformed into the \gls{pig} $\Psi_{\sigma_{1}}$, as depicted in Figure~\ref{fig:pig_example}.

Furthermore, the GGNN model $\mathcal{M}$ requires \glspl{pig}, where each input edge has a discrete edge type assigned to it~\cite{li2015gated}. 
Thus, $\forall ( f_{v}(a_{i}), f_{v}(a_{j})) \in E^{\Psi}_{\sigma_{c}}$ of an \gls{pig} $\Psi_{\sigma_{c}}$ (where $0 < i \leq j \leq \vert \sigma_{c} \vert$), we define the edge type annotation function $f_{et}((x_{1}, x_{2}), \Psi_{\sigma_{c}})$, as formalised in Eq.~(\ref{eq:fkt_anno}).   
\begin{equation}
\label{eq:fkt_anno}
\begin{split}
    f_{et}((x_{1},x_{2}), E^{\Psi}_{\sigma_{c}}) = 
    \begin{cases}
    	\langle\text{Recursive}\rangle & \text{if } x_{1} = x_{2}, \\
    	\langle\text{Start}\rangle & \text{if } x_1 =f_{v}(\langle\text{Start/End}\rangle), \\
    	\langle\text{End}\rangle & \text{if } f_{v}(\langle\text{Start/End}\rangle) = x_2, \\
    	\langle\text{Backward}\rangle & 
    	    \text{if } \exists (x_2, x_1) \in E^{\Psi}_{\sigma_{c}}, \\
    	\langle\text{Forward}\rangle  & 
    	    \text{else}.
    \end{cases}
\end{split}
\end{equation}
The edge type is also stored in the respective edges of $E^{\Psi}_{\sigma_{c}}$. For example, following the insertion of the edge type $\langle\text{Start}\rangle$ between the source and target node, the edge can be represented as $(\langle\text{Start/End}\rangle \text{,} \langle\text{Start}\rangle \text{,} \langle\text{Start Trip}\rangle)$ in $E^{\Psi}_{\sigma_{1}}$.
Figure~\ref{fig:pig_example_vis} depicts the \gls{pig} of our running example $\Psi_{\sigma_{1}}$, including its edge types. 

\input{img/fig_example_pig_vis}

In the last step of event log transformation, we numerically encode the categorical node label (i.e., activity) and edge label values of the \glspl{pig}. The GGNN used in this paper requires a numerical encoding of the input data for calculating forward- and backward-propagation~\citep{li2015gated}. 
To ensure this, we one-hot encode the categorical label values of the \glspl{pig}' nodes and edges (i.e., source node, edge type, and target node). 

\subsection{GGNN Model Creation and Training}
\magic{} uses a GGNN to create and train the model $\mathcal{M}$ for process outcome prediction, using the set of \glspl{pig} $\mathcal{I}$. From the created model, activity relevance scores for individual \glspl{pig} are determined during prediction.
We select a GGNN model because it can directly manage the graph-oriented structure of process data; expressed otherwise, it can explicitly map process activities of \glspl{pig} as nodes and even the relationships between these process activities as edge types and directions. Typically, other ML or DL algorithms are incapable of understanding the semantics of a process to the same extent. This is because they do not encode node and edge information separately from each other, and some neglect edge information entirely \cite{zhou.2018}. Therefore, GGNN models are a promising candidate for capturing process semantics. 

For our GGNN architecture, we used an adapted version of the architecture proposed in \citet{li2015gated}. Their GGNN extends the \textquote{vanilla} GNN of \citet{scarselli2008graph} through using GRUs \citep{cho2014learning} and backpropagation through time (BPTT) \citep{rumelhart.1986} for parameter learning. 
GRUs resolve the problem of gradient vanishing~\citep{bengio.1994}, which occurs when performing backpropagation in GNNs for longer sequences~\cite{li2015gated}. Generally, event logs include several sequences exceeding $50$ steps (cf., the event logs we use in this paper).
On the other hand, the BPTT gradient-based technique enables us to learn the internal parameters of a GGNN more efficiently~\cite{zhou.2018}. Such efficient computation is necessary because event logs can contain several million events.

According to the MPNN framework of~\citet{gilmer2017neural}, the architecture of the GGNN model can be described in terms of message-passing and readout phases. 
The message-passing phase receives as its input \glspl{pig} of $\mathcal{I}$ and returns abstract node representations. In our case, an \gls{pig}’s nodes represent process activities. 
In the message-passing phase, these node representations $\mathbf{h}^{(t+1)}$ are calculated via two steps.
First, for a node $v$, it calculates messages $\mathbf{m}^{(t+1)}_{v}$ by applying the message function $M_t$, as formalised in Eq. (\ref{eq:av}).
\begin{equation}
    \label{eq:av}
     \mathbf{m}_{v}^{(t+1)} = \sum_{w \in f_{nbr}(v)} M_t(\mathbf{h}_v^{(t)}, \mathbf{h}_w^{(t)}, \mathbf{e}_{(v,w)})=\sum_{w \in f_{nbr}(v)}\mathbf{A}_{e(v,w)} \mathbf{h}^{(t)}_w.
\end{equation}

Messages express the interactions between nodes; here, these are the interaction between process activities of \glspl{pig}.   
Given these messages $\mathbf{m}^{(t+1)}_{v}$ and the node representations $\mathbf{h}_{v}^{(t)}$ of node $v$ at time $(t)$, the new node representation $\mathbf{h}_{v}^{(t+1)}$ of node $v$ at time $(t+1)$ can be calculated by using the node update function $f_{GRU}$, as shown in Eq. ~(\ref{eq:au}). 
\begin{equation}
\label{eq:au}
    \textbf{h}_v^{(t+1)} = f_{GRU}(\textbf{h}_v^{(t)},\textbf{m}_v^{(t+1)}).
\end{equation}

For every node of an \gls{pig} $\Psi_{\sigma_{c}} \in \mathcal{I}$, node representations are updated roughly simultaneously for each time step (t). 
Depending on the number of hidden layers and propagation steps per each hidden layer, the update of the node representations is repeated. After completing the last propagation step in the final layer, the message-passing phase outputs the final node representations to the readout phase. 

Then, the readout phase takes as its input the abstract node representations $\mathbf{h}_{v}^{(T)}$ and maps these to a predicted process outcome $\hat{\mathbf{o}}$ via the readout function $R$, as formalised in Eq.~(\ref{eq:ar}): 
\begin{equation}
\label{eq:ar}
    \hat{\mathbf{o}} = R(\{\textbf{h}_v^{(T)}| v \in G\}).
\end{equation}
The predicted process outcome $\hat{\mathbf{o}}$ is a real value lying within the range $[0,1]$.
To learn the GGNN's internal parameters, the loss function \emph{mean squared error} is applied to each data point (i.e., prediction and label) of a batch of \glspl{pig} $\in \mathcal{I}$ and measures the penalty. 
Additionally, a cost function calculates the sum of loss functions over the batch of \glspl{pig} $\in \mathcal{I}$. 
After parameter learning, the values of the GGNN model $\mathcal{M}$ are adjusted.

\subsection{Prediction and Relevance Determining}
\label{sec:predanddeter}
\magic{} extracts the relevance scores for the process activities of an \gls{pig} $\Psi_{\sigma_{c}}\in \mathcal{I}$ from the created GGNN model $\mathcal{M}$ during outcome prediction. 
Given an \gls{pig} $\Psi_{\sigma_{c}}$, the GGNN model $\mathcal{M}$ returns a real-valued process outcome prediction $\hat{\mathbf{o}}$.
If the value of the prediction is $\geq 0.5$, we map the value to $1$; otherwise, we map it to ${0}$. 
During prediction, relevance scores are calculated by the readout phase of the model $\mathcal{M}$, based on the final node representations provided by the message-passing phase. 
The relevance scores are the weights of the nodes representing process activities of an \gls{pig} $\Psi_{\sigma_{c}}$. Such weights express a process activity's importance with regard to a predicted process outcome.
To understand how the relevance scores are calculated in the model $\mathcal{M}$, the readout function $R$ (cf., Eq. (\ref{eq:ar})) can be further described as follows~\cite{li2015gated}:
\begin{equation} 
\label{eq:output} 
    \hat{\mathbf{o}}= R\Bigg(\text{sig}\bigg(\tanh{\Big(\sum_{v \in G} \text{sig}\big(i(\textbf{h}_v^{(T)}, \textbf{h}^0_v)\big) \odot \tanh{\big(j(\textbf{h}_v^{(T)}, \mathbf{h}_v^{0})\big)\Big)\bigg)}}\Bigg),
\end{equation}
where the term $\text{sig}\big(i(\textbf{h}_v^{(T)}, \textbf{h}^0_v)\big)$ calculates the node relevance ${r}_{v}$ for node $v$, and the term $\tanh{\big(j(\textbf{h}_v^{(T)}, \mathbf{h}_v^{0})\big)}$ returns the node representation of node $v$. $i$ and $j$ are neural networks. Both neural networks take as their inputs the concatenation of the final node representation $\mathbf{h}_{v}^{(T)}$ and the initial node representation $\mathbf{h}_{v}^0$. A graph-based representation vector $\mathbf{h}_{G}$ is calculated by point-wise multiplying the output of both terms for each node, calculating the sum over all nodes, and inputting this through a $tanh$ activation function.  
Then, a sigmoid function (sig) is applied to the vector $\mathbf{h}_{G}$, to obtain a process outcome prediction $\hat{\mathbf{o}}$.  

More specifically, the term $\text{sig}\big(i(\textbf{h}_v^{(T)}, \textbf{h}^0_v)\big)$ operates as a soft-attention mechanism; it determines which activities of the \gls{pig} $\Psi_{\sigma_{c}}$ are of greater and lesser relevance to the current graph-based process outcome. The term returns for each node $v$ a real-valued relevance score $r_{v}$. To capture the relevance scores for all activities of the \gls{pig} $\Psi_{\sigma_{c}}$, we store these in a relevance score vector $\mathbf{r}_{\Psi_{\sigma_{c}}}\in \mathbb{R}^{\vert V \vert \times 1}$. Then, we min-max normalise the activities' relevance scores of $\mathbf{r}_{\Psi_{\sigma_{c}}}$. For this, the minimum is set to $0$, and the maximum is set to $1$. Note: The relevance scores of the activities excluded from the \gls{pig} $\Psi_{\sigma_{c}}$ are set to zero. 

For instance, \magic{} determines for the \gls{pig} $\Psi_{\sigma_{1}}$ the relevance score vector $\mathbf{r}_{\Psi_{\sigma_{1}}} = \langle 0.4, 0.1, 0.05, 0.25, 0.1, 0.25 \rangle$, by calculating the process outcome prediction $\hat{\mathbf{o}}_{\Psi_{\sigma_{1}}} = 1$ (\textquote{true}) from the GGNN model $\mathcal{M}$.
In this example, the activity \textquote{Start Trip} obtains the highest relevance score of $0.3$, indicating the importance of this activity for the process outcome (i.e., the overspending of travel expenses).

%% file: tables/tab_eventlog.tex
\begin{table}[htbp]
\caption{Examplary event log $\mathcal{L}_\tau^{ex}$ comprising the trace $\sigma_{1}$.}
\label{tab:log}
\centering
\begin{tabular}{|c|c|c|c|}
\hline
Case & Activiy            & Timestamp         & Travel expense overspent \\ \hline
1    & Start Trip       & 01.02.16 10:06:00 & \multirow{8}{*}{True}                     \\ \cline{1-3}
1    & Permit S    & 01.02.16 11:43:00 &                      \\ \cline{1-3}
1    & Permit A & 01.02.16 13:00:10 &                      \\ \cline{1-3}
1    & Permit A   & 01.02.16 15:10:00 &                      \\ \cline{1-3}
1    & Permit F\_A  & 02.02.16 12:00:04 &                      \\ \cline{1-3}
1    & End trip          & 03.02.16 17:30:39 &                      \\ \cline{1-3}
1    & Send Reminder     & 04.02.16 12:00:00 &                      \\ \cline{1-3}
1    & Send Reminder     & 05.02.16 12:00:00 &                      \\ \hline
\end{tabular}%
\end{table}

%% file: img/fig_example_pig.tex
\begin{figure}[htbp]
    \centering
\footnotesize

\begin{tikzpicture} %
\tikzset{
    vertex/.style={minimum size=1.5em},
    edge/.style={->,> = latex'}
}

\tikzset{
    ncbar angle/.initial=70,
    ncbar/.style={
        to path=(\tikztostart)
        -- ($(\tikztostart)!#1!\pgfkeysvalueof{/tikz/ncbar angle}:(\tikztotarget)$)
        -- ($(\tikztotarget)!($(\tikztostart)!#1!\pgfkeysvalueof{/tikz/ncbar angle}:(\tikztotarget)$)!\pgfkeysvalueof{/tikz/ncbar angle}:(\tikztostart)$)
        -- (\tikztotarget)
    },
    ncbar/.default=0.4cm,
}

\tikzset{round left paren/.style={ncbar=0.4cm,out=150,in=-150}}
\tikzset{round right paren/.style={ncbar=0.5cm,out=60,in=-60}}

\matrix [above right] at (current bounding box.north west) (activities) {
  \node [label=right:$V^{\Psi}_{\sigma_{1}}\text{$=$\{}$] {}; \\
  \node [label=right:$ \langle \text{Start Trip}\rangle\text{,}$] {}; \\
  \node [label=right:$ \langle \text{Permit S}\rangle\text{,}$] {}; \\
  \node [label=right:$ \langle \text{Permit A}\rangle\text{,}$] {}; \\
  \node [label=right:$ \langle \text{Permit F\_A}\rangle\text{,}$] {}; \\
  \node [label=right:$ \langle \text{End trip}\rangle\text{,}$] {}; \\
  \node [label=right:$ \langle \text{Send Reminder}\rangle\text{,}$] {}; \\
  \node [label=right:$ \langle \text{Start/End}\rangle\}$] {}; \\
};

\matrix [ right=.5 em of activities] (edges) {
    \node [label=right:$E^{\Psi}_{\sigma_{1}}\text{$=$\{}$] {}; \\
    \node [label=right:$(\langle\text{Start/End}\rangle \text{,} \langle\text{Start Trip}\rangle)\text{,}$] {}; \\,
    \node [label=right:$(\langle\text{Start Trip}\rangle \text{,} \langle\text{Permit S}\rangle)\text{,}$] {}; \\,
    \node [label=right:$(\langle\text{Permit S}\rangle \text{,} \langle\text{Permit A}\rangle)\text{,}$] {}; \\,
    
    \node [label=right:$(\langle\text{Permit A}\rangle \text{,} \langle\text{Permit A}\rangle)\text{,}$] {}; \\,
    \node [label=right:$(\langle\text{Permit A}\rangle \text{,} \langle\text{Permit F\_A}\rangle)\text{,}$] {}; \\,
    \node [label=right:$(\langle\text{Permit F\_A}\rangle \text{,} \langle\text{End trip}\rangle)\text{,}$] {}; \\,
    \node [label=right:$(\langle\text{End trip}\rangle \text{,} \langle\text{Send Reminder}\rangle)\text{,}$] {}; \\,
    \node [label=right:$(\langle\text{Send Reminder}\rangle \text{,} \langle\text{Send Reminder}\rangle)\text{,}$] {}; \\,
    \node [label=right:$(\langle\text{Send Reminder}\rangle \text{,} \langle\text{Start/End}\rangle)\}$] {}; \\  
};

\draw[-,decorate,decoration={calligraphic straight parenthesis, amplitude=10pt,mirror}] let   \p1=(edges.north),\p2=(edges.south),\p3=(activities.west) in (\x3,\y1) -- (\x3,\y2 ) node [midway,xshift=-0.4cm,left] {$\Psi_{\sigma_{1}}\text{$=$}$};

\draw[-,decorate,decoration={calligraphic straight parenthesis,amplitude=10pt}]
(edges.north east) -- (edges.south east) node [midway,xshift=0.4cm,right] {};

\draw[-,decorate,decoration={text along path,text={|\huge|,},text align={center}}]
(activities) -- (edges);

\end{tikzpicture}

    \caption{\gls{pig} $\Psi_{\sigma_{1}}$ extracted from trace $\sigma_{1}$.}
    \label{fig:pig_example}
\end{figure}

%% file: img/fig_example_pig_vis.tex
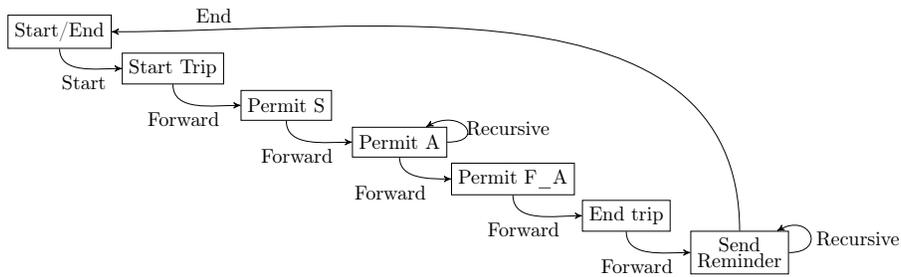
\begin{figure}[h]
    \centering
    
\resizebox{\textwidth}{!}{    
\begin{tikzpicture}[->,>=stealth',auto,node distance=3cm, main node/.style={circle,draw},every text node part/.style={align=center}]
\tikzset{
    vertex/.style={minimum size=1.5em},
    edge/.style={->,> = latex'}
}

\node[vertex, draw] (8) at (-5,0.4) {Start/End};
\node[vertex, draw] (1) at (-3,-0.25) {Start Trip};
\node[vertex, draw] (2) at (-1,-0.9) {Permit S};
\node[vertex, draw] (3) at (1,-1.55) {Permit A};
\node[vertex, draw] (4) at (3,-2.2) {Permit F\_A}; %
\node[vertex, draw] (6) at (5,-2.85) {End trip};
\node[vertex, draw, execute at begin node=\setlength{\baselineskip}{1ex}] (7) at (7,-3.5) {Send\\ Reminder} ;

\draw[] (3) edge[out=0,in=30,looseness=4] (3) node[above right, xshift=3em] (33) {Recursive};
\draw[] (7) edge[out=0,in=30,looseness=4] (7) node[above right, xshift=3.5em] (33) {Recursive};
\draw[] (7) edge[out=90,in=0,looseness=1] (8) node[below left, xshift=-3em] (78) {Forward};
\draw[] (8) edge[out=270,in=180,looseness=1] (1) node[below left, xshift=9em, yshift=1.5em] (78) {End};
\draw[] (1) edge[out=270,in=180,looseness=1] (2) node[below left, xshift=-3em] (78) {Start};
\draw[] (2) edge[out=270,in=180,looseness=1] (3) node[below left, xshift=-3em] (78) {Forward};
\draw[] (3) edge[out=270,in=180,looseness=1] (4) node[below left, xshift=-3em] (78) {Forward};
\draw[] (4) edge[out=270,in=180,looseness=1] (6) node[below left, xshift=-4em] (78) {Forward};
\draw[] (6) edge[out=270,in=180,looseness=1] (7) node[below left, xshift=-3em] (78) {Forward};

\end{tikzpicture}
}
    \caption{Graph-oriented representation of \gls{pig} $\Psi_{\sigma_{1}}$ with edge types.}
    \label{fig:pig_example_vis}
\end{figure}

%% file: sections/05_evaluation.tex
\section{Evaluation}
\label{sec:eval}

\subsection{Procedure}
The goal of the evaluation is to assess (1) the efficacy of our technique and (2) its effectiveness \citep{march1995design}. We consider \magic{} to be efficacious if it delivers relevance scores for process activities with a high faithfulness \citep{du2019techniques}. 
Therefore, we evaluate the predictive quality (which determines the quality of the relevance scores) of the model and compare it against those of other state-of-the-art techniques. 
Furthermore, we verify the relevance scores by repeating the experiments after removing each instance’s most/least relevant activity from one of the datasets, to observe changes in predictive quality. 
Thus, to evaluate the efficacy of \magic{}, we test the following hypotheses: (1) \magic{} exhibits a similar or superior predictive quality to other state-of-the-art algorithms for outcome prediction and (2) removing activities identified by \magic{} as being most relevant has a stronger negative impact on predictive quality than removing activities that it identifies as least relevant.

We consider \magic{} effective if it fulfils the stated goal of supporting process analysts in improving business processes. More specifically, we aim to close the gap between process discovery and process analysis, using \magic. 
We conduct a case study to evaluate the usefulness of the relevance scores determined through \magic{} for process analysts, in terms of identifying the root causes of process performance issues in the process flow.

\subsection{Setup}
To improve model generalisability, we randomly shuffle the process instances of each event log. For this, we perform a process-instance-based sampling to consider the process-instance-affiliations of event log entries.
For each event log, we perform ten-fold cross-validation. Thus, in every iteration, the event log's process instances are split into a 90\%-training and 10\%-testing set. Additionally, we use 10\% of the training set for validation; this prevents overfitting by implementing early stopping after ten epochs (i.e., learning iterations). 

As a benchmark, we use three state-of-the-art \gls{ml} algorithms for predicting process outcomes: \gls{bilstm} \citep{wang.2019}, Random Forest (RF) \citep{senderovich2017intra}, and XGBoost (XG) \citep{teinemaa2019}. \Citet{hinkka2018classifying} demonstrate that GRUs achieve a comparable predictive quality like \textquote{vanilla} LSTMs while being more efficient. However, efficiency is not within the scope of our evaluation.
We decided to use a \gls{bilstm} as \gls{dnn}-based benchmark, because it can exploit the sequences' control-flow information from both directions and it tends to outperform \textquote{vanilla} LSTMs in terms of predictive quality \citep{wang.2019}.

We measure predictive quality (i.e., efficacy) using the following metrics: \emph{Area under the receiver operating characteristic (ROC) curve} ($AUC_{ROC}$), \emph{Specificity}, and \emph{Sensitivity} \citep{sokolova2009}. 
$AUC_{ROC}$ measures a classifier's ability to avoid false classifications~\citep{sokolova2009}.
A major advantage of the $AUC_{ROC}$ over other popular measures---such as the \emph{Accuracy} (overall correctness of a classifier) or \emph{F1-score} (harmonic mean of \emph{Precision} and \emph{Recall})---is that it remains unbiased for a highly imbalanced class label distribution \citep{Kratsch2020}. In outcome prediction scenarios, the distribution of class labels is typically imbalanced~\citep{teinemaa2019}.
Additionally, we use \emph{Specificity} (true negative rate (TNR) = 1 -- false positive rate (FPR)) 
and \emph{Sensitivity} (true positive rate (TPR)) to measure the classifiers' predictive quality. 
The FPR and \emph{TPR} are mapped onto the ROC curve's horizontal and vertical axes, respectively. Therefore, the \emph{Specificity} and \emph{Sensitivity} allow us to better interpret the $AUC_{ROC}$. 
For significance testing, we perform a Friedman test followed by a Nemenyi test (post hoc) as suggested by \citet{demvsar2006statistical} for each data set and metric.
Finally, to intuit the classifier predictions' robustness, we evaluate the standard deviation over the ten folds for each measurement.

For the second part of the evaluation (i.e., evaluating the effectiveness of \magic{}), we use the best models (in terms of the $AUC_{ROC}$) from the first part of the evaluation, to determine relevance scores for single instances. We use the \gls{dfg} miner in pm4py\footnote{\url{http://pm4py.fit.fraunhofer.de/}} to identify the process from the event log, and we colour the activities according to their relevance. \Glspl{dfg} are a user-friendly visualisation that \textquote{shows which activities can follow another directly} \citep{leemans2019directly}. To visualise multiple instances, we split them by outcome label (because the relevance scores are only useful for the predicted outcome of the instance) into two datasets and aggregate the relevance scores for each by finding the mean value.

\subsection{Data}
\label{sec:logs}
We evaluate \magic{} using four real-life event logs, whose characteristics are summarised in Table~\ref{tab:logs}. Three of them originate from the Business Process Intelligence challenges; the other was provided by a mid-sized German home appliances vendor. 

\begin{description}
\item[bpi2017w~\citep{bpi2017}] contains event data describing the loan application process of a Dutch financial institute. 
We only consider \emph{workflow events}, which are executed by humans.
For the outcome prediction target, we select the attribute \emph{accepted}. Therefore, \magic{} determines the relevancy of process activities with respect to the acceptance or rejection of a loan.

\item[bpi2018al~\citep{bpi2018}] describes the European Union’s application process for German farmers (\emph{Application log}). 
For the outcome prediction target, we select the attribute \emph{rejected}, which is highly imbalanced. Therefore, \magic{} determines the relevancy of process activities with respect to the rejection or acceptance of direct payment applications. 

\item[bpi2020pl~\citep{bpi2020}] describes the reimbursement process at the Eindhoven University of Technology (\emph{Permit log}). %
For the outcome prediction target, we select the attribute \emph{travel expense overspent}. Therefore, \magic{} determines the relevancy of process activities with respect to the adherence or non-adherence to travel budgets.

\item[\serviceprocessdata{}~\citep{repair2020}] represents a customer service process for faulty home appliance devices in need of repair. We collected the dataset and published it along with a documentation as part of this research. The process begins with the creation of the repair order; then, it proceeds through the reception and analysis of the device, extending up to the actual repair and the final return of the device to the customer.
We choose the attribute \emph{customer repair on time} as the outcome prediction target. Therefore, \magic{} determines the relevancy of process activities with respect to the meeting or falling short of service agreements (in terms of repair time) with customers. 
\end{description}

\input{tables/tab_eventlog_characteristics}

To run the experiments, we implemented \magic{} using Python. For reproducibility, the source code, event logs, and results can be found on GitHub\footnote{\url{https://github.com/fau-is/grm}}.

\subsection{Results for Predictive Quality}
Table \ref{tab:results} presents the results (averaged over ten folds) for \magic{} and the baseline techniques. In terms of $AUC_{ROC}$, \magic{} outperforms all three baseline techniques for each dataset (signficantly for bpi2018al). Taking a closer look (by considering \emph{Specificity} and \emph{Sensitivity}), it is seen that \magic{} is consistently significantly superior for the less frequent class (in the bpi2017w event log, the negative class is underrepresented, whereas in the other three logs it is overrepresented). We can see that the more distorted the class of interest is, the better \magic's results are for the weaker class compared to the baseline techniques. This observation accords well with the research in \citet{Kratsch2020}, which found that \gls{dl} techniques outperformed traditional \gls{ml} techniques for imbalanced target variables in process outcome prediction. However, our results show that of the \gls{dl} architectures, \glspl{ggnn} clearly outperform LSTMs.

\input{tables/tab_results}
Meanwhile, \magic{} performs significantly worse on three of four datasets for the more frequent class. While \magic{} still performs reasonably well for some of the datasets (e.g., bpi2018al and \serviceprocessdata), the results also suggest that \magic{} performs poorly for a more frequent class (e.g., bpi2020pl). %

This part of the evaluation did not aim to prove that \magic{} is superior to other state-of-the-art techniques but rather to assure a reasonable predictive quality relative to the baselines. The $AUC_{ROC}$ values for \magic{} are better than those of the baseline techniques for all datasets; thus, we are confident that \magic{} can compete against state-of-the-art \gls{pbpm} techniques. However, when using \magic{} to determine the relevance scores for the more frequent class, the predictive quality (sensitivity or specificity)---operating as a proxy for the faithfulness of the relevance scores---must first be assured by the process analyst.

To further substantiate the validity of the relevance scores, we created two new datasets from the \serviceprocessdata{} dataset, by removing the least and most frequent activity from each instance, respectively. The $AUC_{ROC}$ results in Table \ref{tab:results_mostleast} confirm the hypothesis that removing an activity results in a lower predictive quality (i.e., less information for the model). 
More importantly, it confirms our hypothesis that the effect of removing the most relevant activity is significant (for $AUC_{ROC}$ and \emph{Specificity}). Removing the least relevant activity from an instance has little impact on the $AUC_{ROC}$, and the \emph{Sensitivity} even improves slightly. There is a noticeable difference for \emph{Specificity}; however, this did not prove to be significant.

\input{tables/tab_results_most_least}

%% file: tables/tab_eventlog_characteristics.tex
\begin{table}[!htpb]
\caption{Event log characteristics.}
\label{tab:logs}
\begin{threeparttable}
\begin{adjustbox}{max width=\textwidth}
\begin{tabular}{|c|c|c|c|c|c|c|c|}
\hline
Event log & \# instances & \# events & \# activities & \begin{tabular}[c]{@{}c@{}}Target \\ variable\end{tabular}          & \begin{tabular}[c]{@{}c@{}}Class \\ distribution \\ (positive / negative)\end{tabular} \\ \hline
bpi2017w  &    31,500          &          128,227        &        8       & \begin{tabular}[c]{@{}c@{}}Loan accepted\end{tabular} &  73.04 - 26.96                                                           \\ \hline
bpi2018al   & 43,809                                                                 & 2,514,266 & 41            & Rejected                                                            & 0.62 - 99.38                                                  \\ \hline
bpi2020pl & 7,065                                                                & 69,193    & 48            & Overspent                                                           & 26.80 - 73.19                                                 \\ \hline
sp2020                                      &    23,906                                                                & 178,078   & 13            & \begin{tabular}[c]{@{}c@{}}Repair in time\end{tabular}            & 34.65 - 65.35                                                   \\ \hline
\end{tabular}%
\end{adjustbox}
\end{threeparttable}
\end{table}

%% file: tables/tab_results.tex
\begin{table}[!htb]	
\caption{Predictive Quality of \magic.}	
\label{tab:results}	
\begin{threeparttable}	
\small
\begin{tabular}{|c|c|c|c|c|c|}	
\hline	
Event log                  & Technique & $AUC_{ROC}$       & \emph{Sensitivity} & \emph{Specificity}   \\ \hline	
	
\multirow{4}{*}{bpi2017w}  & \magic{}       & \textbf{0.600 (0.007)} & 0.806 (0.057)     & \textbf{0.395** (0.060)} 	\\ \cline{2-5}
                           & BiLSTM    & 0.593 (0.005)          & 0.948 (0.004)              & 0.237 (0.011) 	\\ \cline{2-5}
                           & RF    & 0.589 (0.008)          & 0.941 (0.006)              & 0.238 (0.014) 	\\ \cline{2-5}
                           & XG    & 0.590 (0.005)          & 0.942 (0.004)              & 0.237 (0.010) 	\\ \hline
\multirow{4}{*}{bpi2018al}  & \magic{}       & \textbf{0.942 (0.015)} & \textbf{0.945** (0.032)}     & 0.939 (0.010) 	\\ \cline{2-5}
                           & BiLSTM    & 0.784 (0.19)          & 0.569 (0.401)              & 0.999 (0.000) 	\\ \cline{2-5}
                           & RF    & 0.530 (0.014)          & 0.060 (0.030)              & 1.000 (0.000) 	\\ \cline{2-5}
                           & XG    & 0.546 (0.028)          & 0.094 (0.059)              & 0.999 (0.000) 	\\ \hline
\multirow{4}{*}{bpi2020pl}  & \magic{}       & \textbf{0.625** (0.021)} & \textbf{0.793** (0.060)}     & 0.457* (0.052) 	\\ \cline{2-5}
                           & BiLSTM    & 0.528 (0.017)          & 0.078 (0.045)              & 0.978 (0.011) 	\\ \cline{2-5}
                           & RF    & 0.535 (0.011)          & 0.128 (0.028)              & 0.942 (0.013) 	\\ \cline{2-5}
                           & XG    & 0.541 (0.012)          & 0.138 (0.030)              & 0.943 (0.013) 	\\ \hline
\multirow{4}{*}{\serviceprocessdata}  & \magic{}       & \textbf{0.778 (0.007)} & \textbf{0.813** (0.019)}     & 0.744** (0.014) 	\\ \cline{2-5}
                           & BiLSTM    & 0.763 (0.012)          & 0.645 (0.038)              & 0.881 (0.016) 	\\ \cline{2-5}
                           & RF    & 0.757 (0.006)          & 0.631 (0.016)              & 0.882 (0.009) 	\\ \cline{2-5}
                           & XG    & 0.759 (0.005)          & 0.633 (0.009)              & 0.884 (0.009) 	\\ \hline
\end{tabular}	
\begin{tablenotes}	
\scriptsize	
\item As all Friedman tests indicated a difference between the techniques, we only report the results of the Nemenyi post hoc test.
*/** indicates that a technique was significantly different (i.e. better or worse) from \textbf{all} alternatives. See GitHub repository for detailed results.
\item \textsuperscript{*} $p<0.1$, \textsuperscript{**} $p<0.05$.
\end{tablenotes}	
\end{threeparttable}	
\end{table}	

%% file: tables/tab_results_most_least.tex
\begin{table}[htpb]
\caption{Predictive quality of \magic{} after removing least/most relevant activities.}
\label{tab:results_mostleast}
\begin{threeparttable}
\small
\begin{tabular}{|c|c|c|c|}
\hline
Event log                   & $AUC_{ROC}$       & \emph{Sensitivity} & \emph{Specificity}   \\ \hline
\serviceprocessdata{} &
0.778 (0.007) &
0.813 (0.019) &
0.744 (0.014) 
\\ \hline
\multicolumn{1}{|c|}{\begin{tabular}[c]{@{}c@{}}\serviceprocessdata{}\\ (w/o least relevant) \end{tabular}} &
0.774 (0.006) &
0.818 (0.019) &
0.729 (0.017) 
\\ \hline
\multicolumn{1}{|c|}{\begin{tabular}[c]{@{}c@{}}\serviceprocessdata{}\\ (w/o most relevant) \end{tabular}} & \textbf{0.764* (0.008)} &
\textbf{0.804 (0.018)} &
\textbf{0.724* (0.006)}
\\ \hline
\end{tabular}
\begin{tablenotes}
 \scriptsize
 \item The Friedman test indicated no difference between the \emph{Sensitivity} values but for $AUC_{ROC}$ and \emph{Specificity}. The Nemenyi test showed that \emph{w/o most} is significantly worse than the \emph{original} event log and \emph{w/o least}. See GitHub repository for detailed results.
\item \textsuperscript{*} $p<0.05$.
\end{tablenotes}
\end{threeparttable}
\end{table}

%% file: sections/06_case_study.tex
\subsection{Case Study}
\label{sec:case_study}
To evaluate the utility of \magic{}, we conducted a case study. For this, we sought an organisation that was actively engaging in process improvement and had event log data available for the respective processes. The company that provided us with the \emph{\serviceprocessdata{}} event log is a premium supplier for home appliances, who strives for service excellence. Their portfolio comprises roughly 25 products (not considering remakes of device types). Their sales are exclusively performed by retail partners (i.e., no direct sales); however, customer service is primarily delivered by the company itself, giving it strategic value. One of their important target measures is the percentage of service orders fulfilled within five business days.

Several workshops were held, in which we learned about the company, its products, and the customer service process; these workshops included a visit of the repair shop. In return, we introduced them to \gls{pm} and began a data-driven analysis of their customer service process. In a joint effort between the head of customer service (as process owner), process analysts, and customer service agents (as process participants), we implemented \magic{} to identify and analyse the process in terms of delayed repairs.

\begin{sidewaysfigure}
    \centering
    \includegraphics[width=\columnwidth]{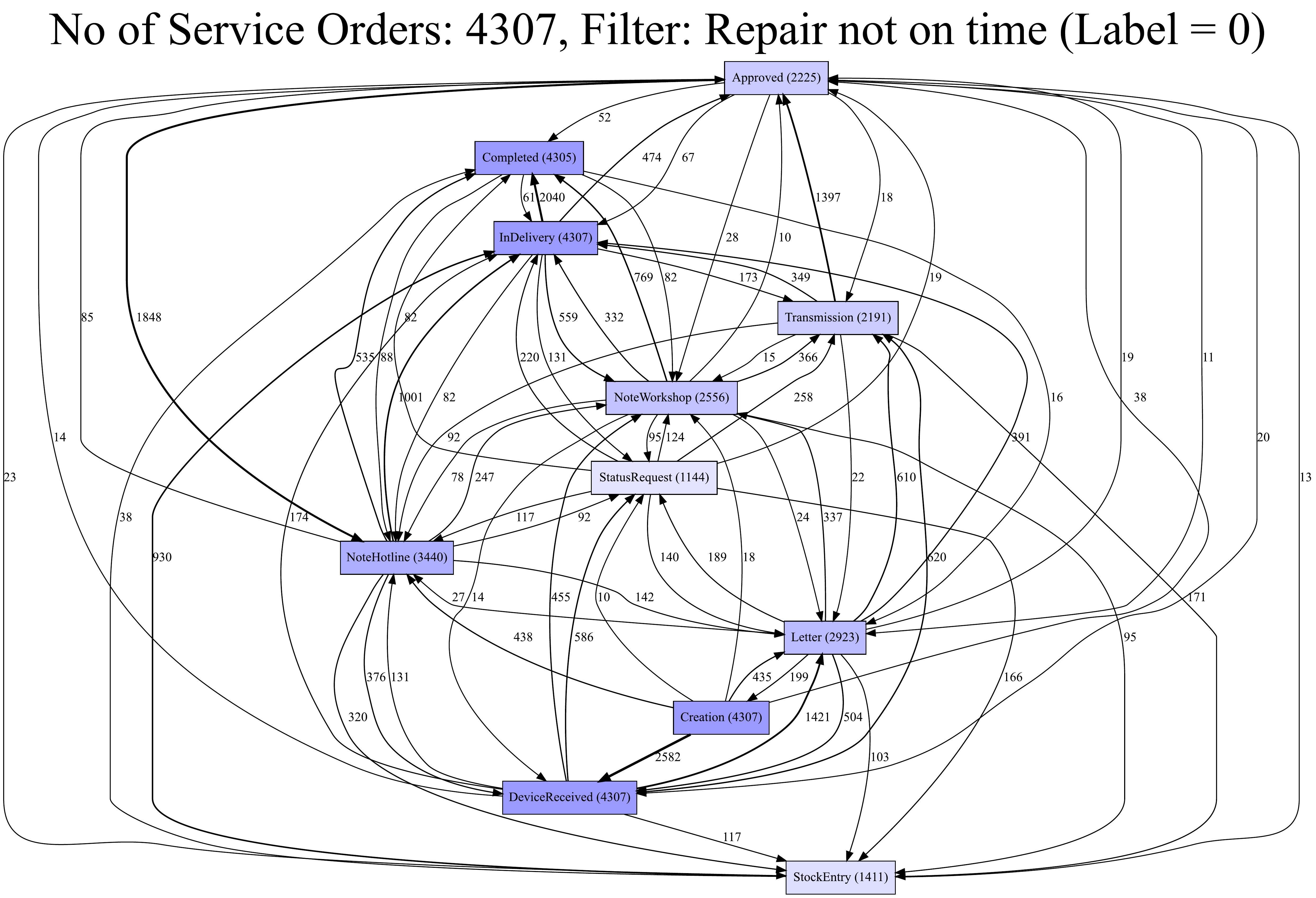}
    \caption{Process model discovered using pm4py's DFG miner with frequency information.}
    \label{fig:sp2020_time_0}
\end{sidewaysfigure}

From the class distribution in Table \ref{tab:logs}, it is evident that only 32.6\% of all repairs could be completed within the desired time-frame of five business days. Hence, the company was eager to improve their repair time and subsequent customer satisfaction. However, their process analysts struggled to identify the root causes for delays within the process execution. Figure \ref{fig:sp2020_time_0} illustrates a \gls{dfg} mined from the \serviceprocessdata{} event log. The frequency was represented by the activities' colours and edge thicknesses (darker blue/thicker = more frequent). This process visualisation represents the current process-discovery capabilities of \gls{pm} software, as we identified from a recent market study\footnote{\url{www.processmining-software.com}}.

While it offered the process analysts some insights pertaining to the repair time (e.g., a high degree of variation was found for non-timely service orders), the analysts struggled to identify root causes for the delays from the process flow. Log filtering was applied as a possible method for isolating the issues, although this was predominantly a tedious procedure of trial and error. 

\begin{sidewaysfigure}
    \centering
    \includegraphics[width=\columnwidth]{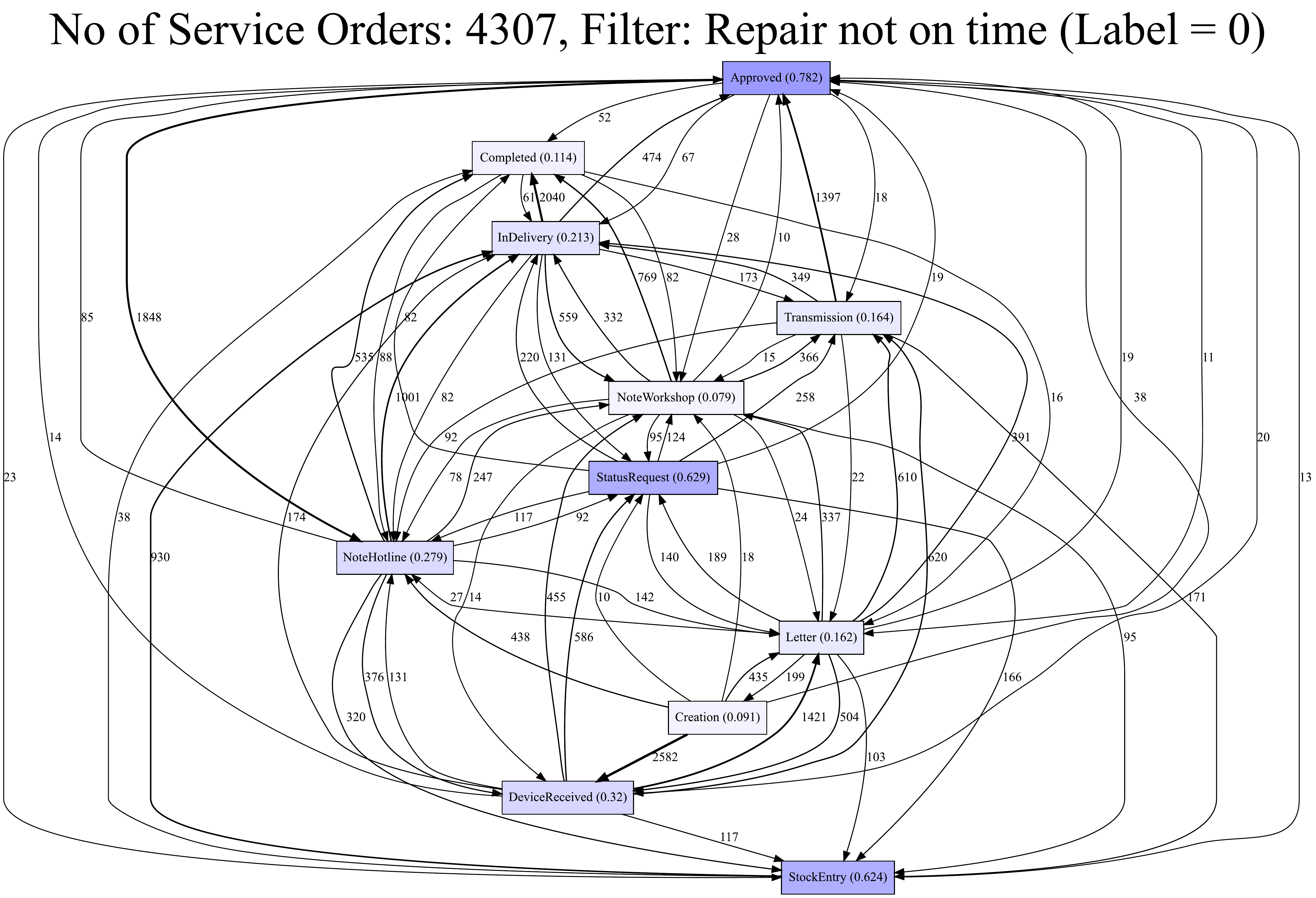}
    \caption{Process model discovered using pm4py's DFG miner augmented by \magic{}'s relevance scores.}
    \label{fig:sp2020_rel_0}
\end{sidewaysfigure}

In contrast, Figure \ref{fig:sp2020_rel_0} shows a \gls{dfg} mined from the \emph{\serviceprocessdata{}} event log, augmented with the relevance scores determined through \magic{}. 
Each process activity is coloured according to its relevance score, which was determined by averaging the scores of all instances with the same outcome prediction (i.e., either positive or negative) contained in the log (darker colours correspond to higher relevance). 

Presented with Figure \ref{fig:sp2020_rel_0} in a workshop, the process analysts were immediately drawn towards the process activity \emph{Approved}. According to the process stakeholders, the activity indicates that the customers were required to provide approval for costs that were incurred for the repair but not covered by the warranty. Further analysis showed that the process was indeed delayed when the activity \emph{Approved} occurred, not only whilst waiting for the approval but also because it occasionally took several days to even request approval from the customer. The company implemented an immediate redesign of the process, by starting low-cost repairs without approval; this was because the risk of losing an unsatisfied customer through long repair times exceeded the risk of bearing the costs. 
Looking at the next most relevant process activities, \emph{StatusRequest} simply indicated that the customer became impatient with the long waiting times, whilst \emph{StockEntry} suggested that missing spare parts delayed the process. Here, an immediate action was to increase the stock for all service points.

To summarise, 
both figures provided insights regarding the delays in repair time. However, the process analysts found it easier to analyse the process model that was augmented with relevance scores based on the business goal. The activities marked as more relevant drew their attention and triggered immediate discussions, resulting in process redesign ideas.

%% file: sections/07_discussion.tex
\section{Discussion}
\label{sec:discussion}

Process analysis---in particular, root cause analysis---is a challenging task and a significant endeavour for organisations, owing to the continuous need to improve business processes for lasting competitiveness \citep{beverungen2020seven}. Manual analysis of a process can be costly and time-consuming. \Gls{pm} has emerged as a data-driven technology to support process analysts. By definition, process discovery in \gls{bpm} facilitates the identification of the as-is process (model) of an organisation \citep[p.155]{Dumas2013}, which is therefore the objective of discovery techniques in \gls{pm}. The discovered process model encourages data analysis from a process perspective, rather than---for example---tables or column charts, which omit the process dimension behind the data. As such, process models are an excellent starting point for process analysis. However, decision support systems in \gls{bpm} must guide process analysts even further in their search for performance issues such as bottlenecks or rework.

Existing studies on process model-based analysis have tried to incorporate this aspect.
\citet{seeliger2019processexplorer} presented \emph{ProcessExplorer} to suggest similar subsets of the process to the analyst. However, whilst recommendations were shown next to the process model, the process model itself was not enriched.
\citet{mannhardt2018mpe} presented a \emph{multi-perspective process explorer} allowing for the projection of performance statistics onto the process model. However, the performance statistics solely relied upon frequency and were not learned.
An example of process model-based analysis was presented by \citet{vanEck2019}, who designed an extension of their \emph{composite state machine miner}, a process discovery technique. They coloured the process nodes according to their degree of artefact interaction; that is, the (\textquote{correlations between sets of artefact states or transitions} \citep{vanEck2019}). However, their technique did not directly permit root cause analysis with respect to performance indicators. Furthermore, the authors stated a limitation of their study: they did not evaluate their work with domain experts. We provide this evaluation via our case study.

To bridge the gap between process discovery and process analysis, we developed \magic{} as a process model-based analysis technique, to identify the relevance of process activities with respect to a business goal (i.e., a process outcome). The quantitative evaluation of \magic{} ensures trust in the validity of the relevance scores. \magic{} provides a reasonable predictive quality, because it can compete against state-of-the-art techniques for process outcome prediction. We were able to demonstrate the impact of process activities that were identified as more relevant on the predictive quality of the model. As such, we find that using a \gls{gnn} is not only beneficial because of its ability to reveal relevance scores but also because it provides good predictive quality; in particular, for imbalanced classification problems.
Furthermore, we evaluated \magic{} via a case study; the results suggest that the relevance scores can help process analysts identify root causes in the process flow and address performance issues.

Besides these contributions, this work also features several limitations. 
First, we argued that business process outcomes are imbalanced and that the \emph{problematic} outcome is typically less frequent. While the case study showed that a violation of this assumption does not necessarily impact the utility of the relevance scores, it remains an aspect that should be carefully evaluated when applying \magic{}.
Second, we did not further evaluate the impact of incorrect predictions on the relevance scores. Currently, we consider the relevance scores of an instance in the context of the predicted label, and this label may be incorrect. %
Third, \glspl{ggnn} are computationally expensive to train. We did not accurately evaluate efficiency; however, from the run-times it was evident that the experiments for the baseline approaches (especially RF) ran significantly faster than those conducted with \magic. 
As run-time correlates with the model's size, especially event logs with a large number of different activities will be challenging to handle. Event logs retrieved from current information systems, such as those used in the evaluation, commonly have a manageable amount of activities. The possible increase in event data in the future---e.g., through sensors and improved logging capabilities---might require a re-evaluation of \magic. 
Whilst this does not impair the theoretical contributions of our work, it may hinder its adoption in practice. 

We see several directions for future research. We presented \magic{} as a technique to close the gap between process discovery and process analysis. However, we believe that \glspl{gnn} can be of use in other phases of the \gls{bpm} life-cycle. Sparse event logs with many different activities pose a challenge for process discovery algorithms in \gls{pm} as models tend to be complex. The relevance scores could be used as an alternative to common threshold parameters for simplifying discovered process models such as fitness \citep{buijs2014quality} or probability \citep{breuker2016comprehensible}.
Especially for processes with many different activities, users could filter on those that have been found most relevant by \magic{} in regards to the defined business goal.
Furthermore, the relevance scores could be used as inputs for some of the redesign heuristics proposed by \citet{reijers2005best}. For example, the heuristic \emph{task elimination} suggests unnecessary tasks that can be removed from a business process. \magic{} could indicate irrelevant process activities with respect to a defined process goal. In another work, we have shown how \glspl{gnn} can provide explainability for predictions in the (predictive) monitoring phase \citep{harl2020ggnn} of the \gls{bpm} life-cycle.

We proposed \magic{} as a technique capable of capturing the semantics of a business process. However, we think that this capability of \gls{gnn} could be exploited even further. Future work should consider using a formal process model such as a Petri net (instead of a graph) as input. This would allow analysts to determine not only relevance scores for activities but also---for example---transitions that represent decision points in the process. Following this line of thought, \gls{gnn} could also be used in combination with decision models \citep{HASIC20181}.

Finally, we plan to extend \magic{} by considering the relevance of contextual attributes of events and process instances alongside the process flow. Contextual information can have a valuable contribution to predictive models~\citep{marquez2017pbpm,brunk2020exploring,BRUNK2021101635}. Potential challenges include incorporating context into the \gls{ggnn} architecture and visualising the contextual attributes in the process model.

%% file: sections/08_conclusion.tex
\section{Conclusion}
\label{sec:conclusion}

In this paper, we presented \magic{}, a GNN-based technique for determining activity relevance scores.
Our work has important implications for both research
and business applications. 
In term of the academic community, our work is an example of applying \gls{ml} to \gls{pm}, to produce results on a process-model level rather than on an instance one \citep{PARK2019113191}. While the latter approach might be suitable for supporting operations, process analysts and managers require actionable insights to achieve long-term improvements for business processes \citep{PARK2019113191}.

We present a novel method with a problem-focused approach. Our technique requires a business goal to be specified in the form of a performance measure. The model learns towards this goal rather than using heuristics (e.g., frequency, similarity, or distance measures); such techniques can provide more guidance for a process analyst considering a business problem \citep{vanEck2019}.

We envisage major implications for business practice. \Gls{pm} is rapidly gaining momentum in practice. Davenport recently suggested that it might even trigger \textquote{a new era of process management} \citep{davenport_2020}. Most commercial \gls{pm} vendors seek to bridge the gap between process discovery and analysis, by adding business intelligence capabilities to their solutions\footnote{\url{www.processmining-software.com}}. Several solutions also offer root cause analysis, combined with the deviations identified through conformance checking. 
Our work highlights the potential of process model-based analysis. Placing the discovered process model at the centre of the analysis facilitates a \emph{process-aware} analysis of the data. %